\documentclass[11pt,a4paper]{article}
\usepackage[hyperref]{emnlp2020}
\usepackage{times}
\usepackage{graphicx}
\usepackage{multirow}
\usepackage{latexsym}
\usepackage{booktabs}
\usepackage{array}

\usepackage{graphicx}
\usepackage[linesnumbered,ruled,vlined]{algorithm2e}
\usepackage{url}
\usepackage{framed}
\usepackage{wrapfig}
\usepackage{tablefootnote}
\usepackage{amsmath}

\usepackage{microtype}
\usepackage{hyperref}
\aclfinalcopy %
\def\aclpaperid{3006} %

\newcommand*{\affaddr}[1]{#1} %
\newcommand*{\affmark}[1][*]{\textsuperscript{#1}}
\newcommand*{\email}[1]{\texttt{#1}}

\title{\wiki: A New Benchmark Dataset for Cross-Lingual Abstractive Summarization}

\author{%
Faisal Ladhak\affmark[1]\Thanks{  Equal contribution.} , Esin Durmus\affmark[2]\footnotemark[1] , Claire Cardie\affmark[2], and Kathleen McKeown\affmark[1] \\
\affaddr{\affmark[1]Columbia University, New York, NY}\\
\affaddr{\affmark[2]Cornell University, Ithaca, NY}\\
\email{\{faisal,kathy\}@cs.columbia.edu} \\ \email{\{ed459\}@cornell.edu},
\email{\{cardie\}@cs.cornell.edu}
}
\date{}

\newcommand\wiki{WikiLingua}

\begin{document}
\maketitle
\begin{abstract}
We introduce \wiki, a large-scale, multilingual dataset for the evaluation of cross-lingual abstractive summarization systems. We extract article and summary pairs in $18$ languages from WikiHow\footnote{\href{https://www.wikihow.com/}{https://www.wikihow.com}}\footnote{The data was collected in accordance with the terms and conditions listed on the website.}, a high quality, collaborative resource of how-to guides on a diverse set of topics written by human authors. We create gold-standard article-summary alignments across languages by aligning the images that are used to describe each how-to step in an article.
As a set of baselines for further studies, we evaluate the performance of existing cross-lingual abstractive summarization methods on our dataset.
We further propose a method for direct cross-lingual summarization (i.e., without requiring translation at inference time) by leveraging synthetic data and Neural Machine Translation as a pre-training step. Our method significantly outperforms the baseline approaches, while being more cost efficient 
during inference.
\end{abstract}

\section{Introduction}
Although there has been a tremendous amount of progress in abstractive summarization in recent years, most research has focused on monolingual summarization because of the lack of high quality multilingual resources \cite{Lewis2019BARTDS,control-over-copying:2020}. While there have been a few studies to address the lack of resources for cross-lingual summarization \cite{giannakopoulos-2013-multi,li-etal-2013-multi,elhadad-etal-2013-multi,nguyen-daume-iii-2019-global}, the datasets employed are very limited in size. Scarcity in the availability of data for cross-lingual abstractive summarization can largely be attributed to the difficulty of collecting high-quality, large-scale datasets via crowd-sourcing. It is a costly endeavor, since it requires humans to read, comprehend, condense, and paraphrase entire articles. Moreover, subjectivity in content selection, i.e. identifying the salient points of a given article, only adds to the difficulty of crowd sourcing this task \cite{nguyen-daume-iii-2019-global}.

\begin{figure*}[th]
\centering
    \includegraphics[width=16cm,height=5.6cm]{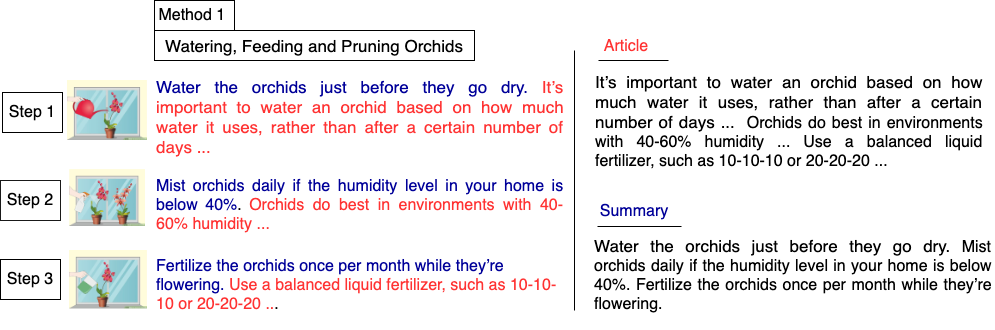}
    \caption{An example method \textit{``Watering, Feeding and Pruning Orchids''} from the guide for \textit{``How to Care for Orchids''}. This method consists of three steps where each step has an illustrative image, a one sentence summary (in blue), and a paragraph providing more details about this step (in red). We combine the paragraphs and summaries from all the steps in each method to create article-summary pairs. 
}
\label{fig:method}
\end{figure*}

To overcome the lack of a large-scale, high quality resource for cross-lingual summarization, we present a new benchmark dataset, \wiki,\footnote{We provide the full dataset, along with the partitions we used in our experiments for this work at: \href{https://github.com/esdurmus/Wikilingua}{https://github.com/esdurmus/Wikilingua}.} which consists of collaboratively written how-to guides with gold-standard summaries across $18$ languages. Each article and summary is written and edited by 23 people, and further reviewed by 16 people, on average, which ensures that the content is of a high-quality. The articles describe multiple methods with steps to complete a procedural task from a diverse set of topics, such as \textit{``How to Make a Creamy Coffee''}, \textit{``How to Exercise to Ease Back Pain''}. Each step contains a one sentence summary followed by a paragraph detailing the instruction, along with an image to illustrate the given instruction, as shown in Figure \ref{fig:method}. Since the ordering of steps may differ for the same article across languages, we align each step using the corresponding illustrative image, as shown in Figure \ref{fig:example}, given that each image is specific to a particular step and shared across languages.\footnote{Some newer, ``in progress" articles do not have images, and in some rare cases an article in one of the languages may use different images. We filter these out.} 

Our final dataset consists of 141,457 unique English articles. Each of the
other 17 languages has, on average, 42,783 articles that align with an article in English.
To the best of our knowledge, \wiki \space is the largest dataset with parallel articles and summaries for cross-lingual abstractive summarization to date. This further opens up avenues to explore new approaches for cross-lingual and multilingual summarization, which are currently understudied.  

With the dataset in hand, we evaluate existing approaches for cross-lingual summarization as baselines. We then propose a method for direct cross-lingual abstractive summarization, leveraging synthetic data and machine translation as a pre-training step. We show that our method outperforms existing baselines, without relying on translation at inference time.

\section{Data Collection and Statistics}
WikiHow is an online resource of how-to guides on a diverse set of topics, written and reviewed by human authors. To ensure high quality content, experts are involved in the writing and reviewing process of these guides.\footnote{\href{https://www.wikihow.com/Experts}{https://www.wikihow.com/Experts}} Each page includes multiple methods for completing a multi-step procedural task along with a one-sentence summary of each step. Figure \ref{fig:method} shows an example method from the guide for  \textit{``How to Care for Orchids''}. For this guide, the method \textit{``Watering, Feeding and Pruning Orchids''} includes three steps. Each step consists of a unique illustrative image, a one sentence summary and a paragraph providing more details. We combine the paragraphs and summaries from all the steps of each method to create article-summary pairs. 
Thus, the summarization task is framed as follows: \textit{given an article detailing instruction on how to complete a procedural task, produce a summary consisting of a list of steps, in the correct order}. This builds on prior work that collected data from WikiHow for monolingual summarization in English \cite{DBLP:journals/corr/abs-1810-09305}. 
We note that, by design, the summaries do not incorporate any potential lead bias, which stands in contrast to single document news summarization, where position is an influential
signal  \cite{BRANDOW1995675}.

\begin{figure*}[th]
\centering
    \includegraphics[width=16cm,height=5.6cm]{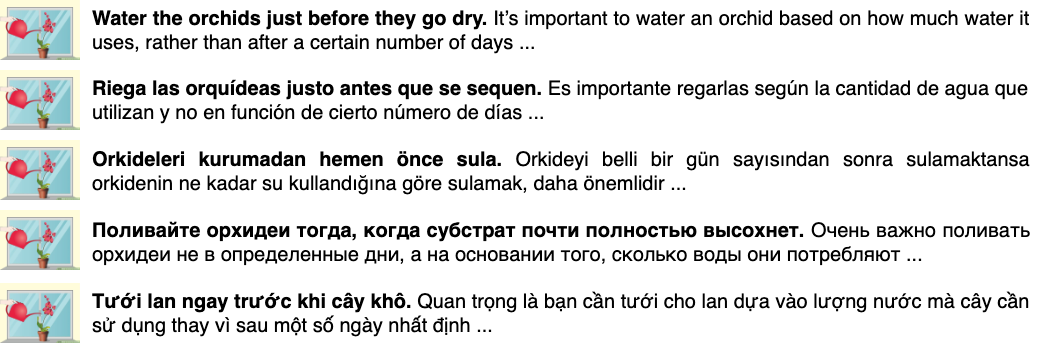}
    \caption{An example step from the guide for \textit{``How to Care for Orchids''}, across five selected languages (top to bottom: English, Spanish, Turkish, Russian and Vietnamese). This shows the summary for the step (bold text), along with the first sentence of the paragraph. Note that the images are the same across the different languages. To get final article-summary pairs, we combine the paragraphs and summaries from all steps in a method. 
}
\label{fig:example}
\end{figure*}

A majority of the non-English guides on this platform are translated from the corresponding English versions by human writers, who are fluent in both English and the target language. Once translated, they are further reviewed by WikiHow's international translation team, before they can be published.
Each of the guides also links to parallel guides in other languages, if available. We collected the guides for all $18$ languages available on WikiHow, and aligned the steps for each method in each guide using the illustrative images. Figure \ref{fig:example} shows an example step from the guide \textit{``How to Care for Orchids''} and its aligned step in five selected languages (English, Spanish, Turkish, Russian, and Vietnamese). This approach ensures that the alignments of the steps are high-quality since the images are unique to each step and shared across all the languages. We merged the step summaries and paragraphs for each WikiHow method as described above to obtain article-summary pairs for all the languages. Table \ref{tab:stats} provides statistics for the number of article-summary pairs in each language that are aligned with articles in English. We note that Turkish, which is the language with the fewest parallel article-summary pairs with English, is still an order of magnitude larger than 
any Langauge in existing cross-lingual datasets.

\begin{table*}[ht]
\centering
\setlength{\tabcolsep}{5pt}
{
\begin{tabular}{lrrrr}
\toprule
& Num. Languages & Num. Summaries & Summary length & Article length\\
&              &   (average)           & (average)      & (average)\\
\toprule
MultiLing'13  & 40 & 30 & 185  & 4,111\\ 
MultiLing'15  & 38 & 30 & 233 & 4,946 \\ 
Global Voices  & 15 & 229 &  51 & 359   \\
\wiki & 18 &  42,783  &  39 & 391 \\

\toprule
\end{tabular}
}
\caption{Comparison of \wiki \space with the existing multilingual summarization datasets. \textit{Num. languages} indicates number of languages covered in each dataset. \textit{Num. summaries} indicates average number of articles per language. \textit{Summary length} and \textit{Article length} corresponds to average number of tokens in summaries and articles respectively.}
\label{tab:comparison_data}
\end{table*}

\begin{table}[ht]
\centering
\setlength{\tabcolsep}{4pt}
{
\begin{tabular}{lcr}
\toprule
Language  & ISO 639-1 & Num. parallel  \\
\toprule
English & en & 141,457 \\
Spanish & es & 113,215\\ 
Portuguese & pt & 81,695 \\ 
French & fr & 63,692\\ 
German & de & 58,375  \\ 
Russian & ru & 52,928 \\ 
Italian & it & 50,968  \\ 
Indonesian & id & 47,511 \\ 
Dutch & nl & 31,270\\ 
Arabic & ar & 29,229\\ 
Chinese & zh & 18,887 \\ 
Vietnamese & vi & 19,600 \\ 
Thai & th & 14,770  \\ 
Japanese & ja & 12,669  \\ 
Korean & ko & 12,189 \\ 
Hindi & hi & 9,929 \\ 
Czech & cs & 7,200 \\ 
Turkish & tr & 4,503\\ 
\toprule
\end{tabular}
}
\caption{Statistics for \wiki. Num. parallel corresponds to the number of articles with a parallel article in English. There are in total 141,457 English article-summary pairs in our dataset.}
\label{tab:stats}
\end{table}

\section{Existing Multilingual Abstractive Summarization Datasets}

There have been a few datasets created for multilingual abstractive summarization tasks in recent years, which we describe in this section.

\textbf{MultiLing'13 and '15.} Multiple versions of the MultiLing dataset have been collected by the organizers of MultiLing Workshops \cite{giannakopoulos-2013-multi,elhadad-etal-2013-multi,kubina-etal-2013-acl}. The MultiLing'13 dataset includes summaries of 30 Wikipedia articles per language, describing a given topic. For MultiLing'15, an additional $30$ documents were collected for evaluation purposes \cite{giannakopoulos-etal-2015-multiling}. We note that while this dataset contains article and summaries in several 
languages
there are no parallel articles or summaries, which makes it difficult to use this dataset for cross-lingual evaluation.

\textbf{Global Voices.} \newcite{nguyen-daume-iii-2019-global} collected social network descriptions of news articles provided by
Global Voices.\footnote{\href{https://globalvoices.org/}{https://globalvoices.org/}} These descriptions, however, are not written with the purpose of summarizing the article content but rather to draw user clicks on social media;
therefore, they have a lower coverage of the original article than a good summary would. 
To address this problem, the authors crowd-source a small set of summaries, in English, for 15 languages. 
We report 
statistics
only on the crowd-sourced summaries, given the click-bait nature of the social media descriptions. Note that unlike our dataset, this one contains summaries only in English, which makes it difficult to evaluate cross-lingual summarization into other languages.

Statistics for the datasets are provided in Table \ref{tab:comparison_data}. \wiki \space is similar to Global Voices in terms of article and summary length while MultiLing articles and summaries are longer. All three existing datasets are limited in size in comparison to \wiki.  Furthermore, our dataset includes articles on a wide-range of topics and the average number of articles per language is two orders of magnitude larger than Global Voices, which is the largest dataset to date for cross-lingual evaluation. The Data Statement \cite{bender-friedman-2018-data} for our dataset can be found in Appendix \ref{data_statements}.

\begin{table}[ht]
\centering
\setlength{\tabcolsep}{5pt}
\scalebox{1}{
\begin{tabular}{lrrr}
\toprule
 &  Train & Validation &  Test  \\ 
\toprule
Spanish & 81,514 & 9,057 & 22,643\\
Russian  & 38,107 & 4,234 & 10,586 \\ 
Vietnamese  & 9,473 & 1,052 &  2,632\\ 
Turkish  &  3,241 & 360 & 901\\ 
\toprule
\end{tabular}
}
\caption{Number of examples in Train/Validation/Test splits per language.}
\label{tab:data_size}
\end{table}

\section{Cross-lingual Experiments}
Following the prior work in cross-lingual abstractive summarization \cite{nguyen-daume-iii-2019-global,ouyang-etal-2019-robust}, we aim to generate English summaries from non-English articles, as an initial study. We experiment with five languages (i.e.\ English, Spanish, Russian, Turkish, and Vietnamese) covering three language families  (i.e.\  Indo-European, Ural-Altaic and Austroasiatic). We split the data for each of the four non-English languages into train/dev/test splits. When splitting the English data, we ensure that all articles from the same topic as  test articles in any of the four non-English languages, are included in the test set. This leaves us with $\sim$ 69K English articles that we randomly split into train and dev set (90/10 split). See Appendix \ref{en_split} for more information.

We use large, pre-trained language models as a starting point for our experiments, given their success on a variety of downstream Natural Language Processing tasks \cite{devlin-etal-2019-bert}, including state of the art results for text summarization \cite{lewis2019bart, liu-lapata-2019-text}. In particular, we use mBART \cite{liu2020multilingual}, which is a multi-lingual language model that has been trained on large, monolingual corpora in 25 languages. The model uses a shared sub-word vocabulary, encoder, and decoder across all 25 languages, and is trained as a denoising auto-encoder during the pre-training step. \newcite{liu2020multilingual} showed that this pre-training method provides a good initialization for downstream machine translation tasks, particularly in lower resources settings, making this an ideal starting point for our cross-lingual summarization experiments. We also ran initial experiments with non-pretrained transformer models, but the results were significantly worse than those with the pre-trained models.

\begin{figure*}[th]
\centering
    \includegraphics[width=16cm,height=2.6cm]{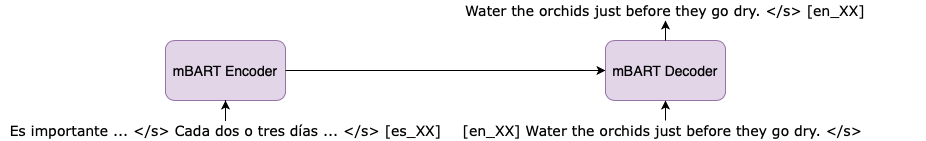}
    \caption{An example showing the fine-tuning procedure for cross-lingual summarization from Spanish to English.}
\label{fig:model}
\end{figure*}

We fine-tune mBART for both monolingual and cross-lingual summarization as a standard sequence-to-sequence model, where the input document is represented as a sequence of tokens (sub-word units), with a special separator token between each sentence, and a language indicator token at the end of the document. The output summary is represented in a similar manner, with a language indicator token at the beginning of the sequence, to prime 
the decoder for generation in the target language, as shown in Figure \ref{fig:model}. We use Fairseq \cite{ott2019fairseq} for all our experiments, and we follow the hyper-parameter settings that were used by \newcite{lewis2019bart} to fine-tune BART for monolingual summarization in English. See Appendix \ref{reproducibility} for more details.

\subsection{Baselines}
We evaluate the following baseline approaches for cross-lingual summarization on our data:

\textbf{lead$_{n}$:} copies first $n$ sentences from the corresponding parallel English source articles. We report results for $n=3$ since it performs the best. 

\textbf{Summarize-then-translate (Sum-Trans):} We fine-tune mBART for monolingual summarization in the source language, and then at inference time, we summarize the article and then translate the summary into the target language. This approach is useful when the source language is higher resource for the summarization task, since it requires translating summaries, which tend 
to
be much shorter than the actual articles, which means fewer opportunities for translation errors. 

\textbf{Translate-then-Summarize (Trans-Sum):} We fine-tune mBART for monolingual summarization in the target language and at inference time, we translate the source language articles into the target language, and then summarize the translation. This approach is useful when the target language is higher resource for the summarization task, though translating entire articles provides more opportunities for translation errors.

\textbf{Trans-Sum-R:} This method, a variation of the translate then summarize method above, first performs a round-trip translation of articles from, and back to, the target language, through the source language, to get noisy articles in the target language. The noisy articles are then paired with the original, clean summary, to train a summarization system in the target language \cite{ouyang-etal-2019-robust}. The summarization system, in this case, can account for potential noise in the translated source article, by learning to generate clean summaries from noisy articles.
For all baselines that require translation, we used the Amazon Web Services (AWS) Translate service, which is among the state of the art Neural Machine Translation systems.\footnote{\href{https://aws.amazon.com/translate/}{https://aws.amazon.com/translate/}}

\textbf{Trans-Sum-G:} This model is the same as the \textbf{Trans-Sum} model except that at inference time, we use the gold translation of the source language article instead of the machine translated one. This is an oracle system that represents the performance we could expect if we had no translation errors. Thus the drop in performance from \textbf{Trans-Sum-G} to \textbf{Trans-Sum} or \textbf{Trans-Sum-R} can be attributed to translation errors.

\begin{table*}[ht]
\centering
\setlength{\tabcolsep}{1.5pt}
\scalebox{1}{
\begin{tabular}{lc||c||c||c}
\toprule
& Es-En & Tr-En & Ru-En & Vi-En \\ 
\toprule
Trans-Sum-G
& $41.66 / 18.64 / 35.07$   

& $45.82 / 22.42 / 39.05$      
& $40.98 / 18.27 / 34.74$
& $41.37 / 18.56 / 35.22$\\ 
\toprule
lead$_{3}$ 
& $24.35 / 06.03 / 16.39$
& $24.55 / 05.98 / 16.49$
& $23.43 / 05.56 / 15.81$
& $22.92 / 05.41 / 15.47$
\\ 
Sum-Trans 
& $36.03 / 13.02 / 29.86$
& $31.57 / 10.45 / 24.76$
& $29.75 / 08.83 / 24.36$
& $26.95 / 07.04 / 21.62$
\\ 
Trans-Sum 
& $37.16 / 14.25 / 31.04$
& $41.06 / 17.72  / 34.53$
& $33.59 / 11.60 / 28.15$
& $34.77 / 12.37 / 29.27$ 
\\
Trans-Sum-R
& $38.13 / 14.95 / 31.96$
& $42.33 / 18.79 / 35.81$
& $34.64 / 12.58 / 29.18$
& $36.29 / 13.21 / 30.57$\\
\toprule
DC
& $38.30 / 15.37 / 32.40^\dag$
& $33.68 / 12.74 / 27.62$
& $32.91 / 11.83 / 27.69$
& $31.89 / 11.07 / 26.36$
\\
DC+Synth
& $40.00 / 16.38 / 33.48^\dag$ 
& $41.76 / 18.84 / 35.78$
& $36.82 / 14.41 / 31.18^\dag$
& $36.48 / 14.29 / 30.96^\ddag$
\\
DC+Synth+MT
& $\textbf{40.60} / \textbf{16.89} / \textbf{34.06}^\dag$
& $\textbf{42.76} / \textbf{20.47} / \textbf{37.09}^\ddag$
& $\textbf{37.09} / \textbf{14.81} / \textbf{31.67}^\dag$
& $\textbf{37.86} / \textbf{15.26} / \textbf{32.33}^\dag$
\\

\toprule
\end{tabular}
}
\caption{Cross-lingual summarization results. The numbers correspond to $\text{ROUGE-1}/\text{ROUGE-2}/\text{ROUGE-L}$ F1 scores respectively. \dag \space indicates where ROUGE-L F1 is significantly better than all baselines, and \ddag \space indicates where ROUGE-L F1 is significantly better than all baselines except Trans-Sum-R. We use Welch’s t-test, and use $p < 0.01$ to assess significance.}
\label{tab:results}
\end{table*}

\subsection{Direct Cross-lingual Summarization}
Most work in cross-lingual summarization has relied on different variations of a two-step approach to cross-lingual summarization, i.e. translation and summarization. Besides the issue of error propagation, another major drawback of such approaches is that they rely on translation at inference time, which makes inference costly as it requires running both a translation system and a summarization system, in sequence. In a real-world scenario, such systems would have a recurring latency and monetary cost for each inference request. Therefore, it is preferable to have cross-lingual summarization methods that do not rely on running an additional translation system at inference time.

The popularity of existing two-step approaches for cross-lingual summarization can largely be attributed to the data that is available -- there are plenty of resources for both machine translation and monolingual English summarization as separate tasks. However, resources that contain parallel articles in multiple languages, with corresponding parallel summaries are scarce. Since our dataset has gold standard translations between English and the other languages, it allows us to explore methods for direct cross-lingual summarization, and measure how they stack up against existing baselines. Furthermore, since we have gold translations, we can directly measure the drop in performance due to translation errors for translate-then-summarize, for each language pair, and see how much of that can be recovered by proposed methods.

For direct cross-lingual summarization, we fine-tune mBART with input articles from the source language, and summaries from the target language \textbf{(DC)}. This setting requires that the model learn both translation and summarization, which requires a large amount of cross-lingual training data. To overcome this, we first propose to generate additional synthetic data by translating the English training articles into the target language \textbf{(DC+Synth)},
using AWS Translate, and pairing them with the original summary in English. Translating training data has been shown to be an effective strategy for cross-lingual transfer for text classification and sequence labeling tasks \cite{Schuster2019CrosslingualTL}.
We note that while this method still relies on machine translation, the cost of translation is shifted to training time, and thus is a 
one-time cost.

Since a cross-lingual summarization model needs to learn how to translate salient information from one language to another, we hypothesize that training the model for machine translation can improve performance of cross-lingual summarization. Therefore, we propose a two-step fine-tuning approach, where we first fine-tune the mBART model for document level machine translation from the source language into English, and then we further fine-tune the model for cross-lingual summarization \textbf{(DC+Synth+MT)}. Similar to above, since we only have a limited amount of parallel document pairs in our dataset, we translate English documents into the source language to create additional parallel data. This method of back-translation to create additional parallel data has been shown to be effective in improving the performance of neural machine translation systems \cite{sennrich-etal-2016-improving, hoang-etal-2018-iterative, edunov-etal-2018-understanding}.\footnote{While back-translation typically uses an intermediate training checkpoint to create synthetic data, we instead use AWS translate.}
\begin{table*}[t]
\centering
\setlength{\tabcolsep}{3pt}
\scalebox{0.9}{
\begin{tabular}{p{15cm}}
\toprule
\textbf{Topic:} How to critique a speech: Assessing the Delivery. \\
\toprule
\textbf{Article:} Does the speaker talk in a way that makes you want to keep listening, or is it easy to tune out? ...  The way the speaker holds him or herself should project confidence and charisma, making the audience feel engaged and included ... Too many “ums”, “likes” and “uhs” take away from a speaker’s credibility ... A great speaker should have memorized the speech long in advance ... Look for signs that the speaker is nervous so you can offer a critique that will help him or her improve next time ...  \\ 
\toprule
\textbf{Reference:} Listen to the speaker’s voice inflections. Watch the speaker’s body language. Listen for filler words. See if the speech was memorized. Assess how the speaker manages anxiety.  \\ 
\toprule
\textbf{Trans-Sum:} Keep the listener's attention. Maintain good posture. Memorize speech beforehand. Identify signs of nervousness.\\
\toprule
\textbf{Trans-Sum-R:} Recognize the listener's needs. Pay attention to the posture of the speaker. Memorize the speech. Recognize the signs of nervousness. \\ 
\toprule
\textbf{DC+Synth+MT (Ours):} Pay attention to the way the speaker is speaking. Notice the way the speaker uses body language. Keep track of the words they say. Remember what they have to say. Watch for signs of nervousness. \\
\toprule
\end{tabular}
}
\caption{An example output summary for Trans-Sum, Trans-Sum-R and DC+Synth+MT. Human annotators preferred the output from DC+Synth+MT.}
\label{tab:example_analysis}
\end{table*}

\section{Results and Analysis}

Table \ref{tab:results} shows ROUGE scores \cite{lin-2004-rouge} for the baselines and proposed cross-lingual approaches. We observe that the lead baseline performs poorly for this task, unlike in the news domain where it's shown to be a strong baseline
\cite{BRANDOW1995675}.

When comparing the performance of Trans-Sum vs.\ Sum-Trans, we find that performance depends on the amount of summarization data available in the source language. Similar to previous work \cite{ouyang-etal-2019-robust}, we find that Tran-Sum works 
significantly better when the amount of data in the source language is limited. However, as source language training data size increases, we see that the gap in performance decreases, as in the case of Spanish, which is similar in size to English, vs.\ Turkish, which is the lowest resource language for summarization in our dataset. This suggests that when the source language data is comparable in size or larger than the target language data, Sum-Trans approach may be worthwhile to consider, as suggested by \newcite{wan-etal-2010-cross}, since it is more cost effective (translating summaries instead of whole articles) and may avoid error propagation from translation systems.

Amongst the baseline methods, Trans-Sum-R works the best. It consistently does better than Trans-Sum baseline, suggesting that round-trip translation to create noisy data can be an effective way to make the model more robust to translation errors at inference time. Since we have gold translations \textbf{(Trans-Sum G)} for each of the articles, we can measure the translation error in the Trans-Sum system. We see that on average, the round-trip translation method is able to recover about 22\% of the performance loss due to translation errors.

For direct cross-lingual summarization, we find that the performance of the base model (DC) is worse than the translate-then-summarize baselines for all languages except Spanish, where it is better. This suggests that direct cross-lingual summarization is a difficult task and requires a larger amount of data, even with a pre-trained mBART model as a starting point. Once we add some synthetic data (DC+Synth), we see the performance improves significantly, especially for the lower resource languages (Tr and Vi), which are on par with the best baseline model. Note that the DC+Synth models would still be preferable, even over the best baseline, as they give similar performance while being much more cost effective for inference. 

Finally, we see that fine-tuning the mBART model for document-level machine translation, before fine-tuning it for cross-lingual summarization, further improves the performance for all languages. This variant (DC+Synth+MT) performs significantly better than all baseline systems for Spanish, Russian and Vietnamese. For Turkish, the performance of DC+Synth+MT is statistically the same as Trans-Sum-R;
we note, however, that our model is significantly better than the Trans-Sum baseline, while the Trans-Sum-R model is not.

\subsection{Human Evaluation}
We ask human annotators on Mechanical Turk to evaluate the generated summaries for fluency and content overlap with the gold reference summary.\footnote{The reference was only shown when evaluating for content overlap, and not for fluency evaluation.} We randomly sample $100$ articles per language and generate summaries using Trans-Sum, Trans-Sum-R and DC+Synth+MT.  Each annotator is shown all three summaries for the same article, along with the reference, and asked to score the summaries for fluency and content on a scale from 1 to 3.
Each of the examples was evaluated by three annotators. To ensure for quality, we filter out annotators with a low agreement score with other annotators who performed the same tasks. The average pairwise agreement between annotators is 56.5\%. 

Table \ref{tab:human_eval} shows that human annotators find all three systems relatively fluent overall. This can be attributed to the use of mBART, which has been pre-trained on large amounts of monolingual data. While there is no significant difference between Trans-Sum-R and DC+Synth+MT, we note that DC+Synth+MT scored significantly higher than Trans-Sum, while Trans-Sum-R is statistically the same as Trans-sum. In terms of content overlap with the reference, we find that DC+Synth+MT model scored significantly better than both the baseline systems (p$\leq$0.05), which validates the ROUGE score improvements we show in Table \ref{tab:results}. Note that the baselines systems are statistically the same in terms of content. Table \ref{tab:example_analysis} shows an example of an article and corresponding output summaries for each of the three systems evaluated. We can see that all the system generated summaries are fluent, however DC+Synth+MT has better overlap with the content in the reference summary.\footnote{More examples are provided in Appendix \ref{appendix_examples}.}

\begin{table}[ht]
\centering
\setlength{\tabcolsep}{3pt}
\scalebox{0.9}{
\begin{tabular}{l r r}
\toprule
Model & Fluency & Content \\
\toprule
Trans-Sum & 2.61 & 2.07 \\ 
Trans-Sum-R & 2.62 & 2.09 \\ 
DC+Synth+MT &  \textbf{2.67} & \textbf{2.19} \\ 
\toprule
\end{tabular}
}
\caption{Human evaluation scores on a scale of 1-3.}
\label{tab:human_eval}
\end{table}

\section{Related Work}

\textbf{Abstractive Summarization.} The majority of research in abstractive summarization has focused on monolingual summarization in English \cite{gehrmann2018bottom, control-over-copying:2020, xsum-emnlp}. \newcite{rush-etal-2015-neural} proposes the first neural abstractive summarization model using an attention-based convolutional neural network encoder and a feed-forward decoder. \newcite{chopra-etal-2016-abstractive} shows improvements over this model using a recurrent neural network for the decoder. \newcite{nallapati-etal-2016-abstractive} shows further improvements by incorporating embeddings for linguistic features such as part-of-speech tags and named-entity tags into their model, as well as a pointer network \cite{NIPS2015_5866} to enable copying words from the source article. \newcite{see-etal-2017-get} extends this model by further incorporating a coverage penalty to address the problem of repetitions in the generated summary. 

\newcite{chen-bansal-2018-fast} takes a two stage approach to abstractive summarization by learning an extractor to select salient sentences from the articles, and an abstractor to rewrite the sentences selected by the extractor. They further train the extractor and abstractor end-to-end with a policy-gradient method, using ROUGE-L F1 as the reward function. Recently, pre-trained language models have achieved the state of the art results in abstractive summarization  \cite{lewis2019bart,liu-lapata-2019-text, control-over-copying:2020}. Therefore, we use mBART \cite{liu2020multilingual} for all the baselines and our direct cross-lingual models.

\textbf{Cross-lingual Abstractive Summarization.} \newcite{wan-etal-2010-cross} proposes summarize-then-translate and translate-then-summarize as approaches for doing cross-lingual summarization. They suggest that summarize-then-translate is preferable because it is computationally less expensive since it translates the summary rather than article, and therefore is less prone to error propagation from translation systems. 
As we show in our work, however, this approach requires a large amount of training data in the source language to build an effective summarization system. On the other hand, translate-then-summarize approach relies on having an accurate translation system and a large amount of summarization training data in the target language. Although translate-then-summarize \cite{10.1145/979872.979877}  and summarize-then-translate \cite{Lim2004MultiDocumentSU,orasan-chiorean-2008-evaluation,wan-etal-2010-cross} are widely used approaches in prior studies, they are prone to error propagation. 
\newcite{ouyang-etal-2019-robust} propose a variant of the translate-then-summarize approach to cross-lingual summarization, by doing a round-trip translation of English articles through the source language to get noisy English articles. They then train on noisy article and clean summary pairs, which allows them to account for potential translation noise.

There is limited prior work in direct cross-lingual summarization. \newcite{10.1109/TASLP.2018.2842432} propose zero-shot cross-lingual headline generation to generate Chinese headlines for English articles, via a teacher-student framework, using two teacher models. 
\newcite{duan-etal-2019-zero} propose a similar approach for cross-lingual abstractive sentence summarization. We note that our approach is much simpler and also focuses on a different kind of summarization task. 

\newcite{zhu-etal-2019-ncls} use round-trip translation of large scale monolingual datasets \cite{10.5555/2969239.2969428,zhu-etal-2018-msmo, hu-etal-2015-lcsts} to generate synthetic training data for their models, and train a multi-task model to to learn both translation and cross-lingual summarization. We tried their approach on our data, using the code provided,\footnote{\href{https://github.com/ZNLP/NCLS-Corpora}{https://github.com/ZNLP/NCLS-Corpora}} 
but the results were worse than all baselines except lead.\footnote{This model gets
ROUGE-L F1 scores of $22.49$, $23.38$, $20.79$, $19.45$ for Spanish, Turkish, Russian and Vietnamese respectively.} We suspect that this may be due to the amount of training data, as their synthetic dataset was much larger than ours (1.69M pairs for Zh-En). An extension of their approach would be to incorporate multi-task training for pre-trained mBART, which we leave for future work. 
Scarcity of cross-lingual summarization data has limited prior work to a few languages, and mostly in the news domain \cite{wan-etal-2010-cross, wan-2011-bilingual, yao-etal-2015-phrase, 7502066, wan2019cross}. While there is some existing work trying to address this
\cite{nguyen-daume-iii-2019-global}, the proposed dataset is still limited in size, and contains summaries only in English. We address this limitation by proposing a new benchmark dataset.

\section{Conclusion}
We present a benchmark dataset for cross-lingual and multilingual abstractive summarization. We then evaluate existing methods in cross-lingual abstractive summarization. We further propose an end-to-end method for direct cross-lingual summarization and show that it achieves significantly better performance than the baselines while being more cost effective for inference. 

Our new benchmark dataset opens up interesting new directions for research in summarization. We would like to further explore multi-source cross-lingual summarization architectures, i.e. models that can summarize from multiple source languages in to a target language. Another interesting avenue would be to explore the feasibility of multilingual summarization, i.e. building models that summarize articles from any language to any other language for a given set of languages.

\section{Acknowledgements}
We would like to thank Chris Kedzie and the anonymous reviewers for their feedback. This research is based on work supported in part by the Office of the Director of National Intelligence (ODNI), Intelligence Advanced Research Projects Activity (IARPA), via contract FA8650-17-C-9117. This work is also supported in part by National Science Foundation (NSF) grant 1815455 and Defense Advanced Research Projects Agency (DARPA) LwLL  FA8750-19-2-0039. The views and conclusions contained herein are those of the authors and should not be interpreted as necessarily representing the official policies or endorsements, either expressed or implied, of ODNI, IARPA, NSF, DARPA or the U.S. Government. The U.S. Government is authorized to reproduce and distribute reprints for governmental purposes notwithstanding any copyright annotation therein.

\bibliographystyle{acl_natbib} 
\bibliography{anthology,emnlp2020}

\begin{thebibliography}{44}
\expandafter\ifx\csname natexlab\endcsname\relax\def\natexlab#1{#1}\fi

\bibitem[{Bender and Friedman(2018)}]{bender-friedman-2018-data}
Emily~M. Bender and Batya Friedman. 2018.
\newblock \href {https://doi.org/10.1162/tacl_a_00041} {Data statements for
  natural language processing: Toward mitigating system bias and enabling
  better science}.
\newblock \emph{Transactions of the Association for Computational Linguistics},
  6:587--604.

\bibitem[{Brandow et~al.(1995)Brandow, Mitze, and Rau}]{BRANDOW1995675}
Ronald Brandow, Karl Mitze, and Lisa~F. Rau. 1995.
\newblock \href {https://doi.org/https://doi.org/10.1016/0306-4573(95)00052-I}
  {Automatic condensation of electronic publications by sentence selection}.
\newblock \emph{Information Processing \& Management}, 31(5):675 -- 685.
\newblock Summarizing Text.

\bibitem[{Chen and Bansal(2018)}]{chen-bansal-2018-fast}
Yen-Chun Chen and Mohit Bansal. 2018.
\newblock \href {https://doi.org/10.18653/v1/P18-1063} {Fast abstractive
  summarization with reinforce-selected sentence rewriting}.
\newblock In \emph{Proceedings of the 56th Annual Meeting of the Association
  for Computational Linguistics (Volume 1: Long Papers)}, pages 675--686,
  Melbourne, Australia. Association for Computational Linguistics.

\bibitem[{Chopra et~al.(2016)Chopra, Auli, and
  Rush}]{chopra-etal-2016-abstractive}
Sumit Chopra, Michael Auli, and Alexander~M. Rush. 2016.
\newblock \href {https://doi.org/10.18653/v1/N16-1012} {Abstractive sentence
  summarization with attentive recurrent neural networks}.
\newblock In \emph{Proceedings of the 2016 Conference of the North {A}merican
  Chapter of the Association for Computational Linguistics: Human Language
  Technologies}, pages 93--98, San Diego, California. Association for
  Computational Linguistics.

\bibitem[{Devlin et~al.(2019)Devlin, Chang, Lee, and
  Toutanova}]{devlin-etal-2019-bert}
Jacob Devlin, Ming-Wei Chang, Kenton Lee, and Kristina Toutanova. 2019.
\newblock \href {https://doi.org/10.18653/v1/N19-1423} {{BERT}: Pre-training of
  deep bidirectional transformers for language understanding}.
\newblock In \emph{Proceedings of the 2019 Conference of the North {A}merican
  Chapter of the Association for Computational Linguistics: Human Language
  Technologies, Volume 1 (Long and Short Papers)}, pages 4171--4186,
  Minneapolis, Minnesota. Association for Computational Linguistics.

\bibitem[{Duan et~al.(2019)Duan, Yin, Zhang, Chen, and
  Luo}]{duan-etal-2019-zero}
Xiangyu Duan, Mingming Yin, Min Zhang, Boxing Chen, and Weihua Luo. 2019.
\newblock \href {https://doi.org/10.18653/v1/P19-1305} {Zero-shot cross-lingual
  abstractive sentence summarization through teaching generation and
  attention}.
\newblock In \emph{Proceedings of the 57th Annual Meeting of the Association
  for Computational Linguistics}, pages 3162--3172, Florence, Italy.
  Association for Computational Linguistics.

\bibitem[{Edunov et~al.(2018)Edunov, Ott, Auli, and
  Grangier}]{edunov-etal-2018-understanding}
Sergey Edunov, Myle Ott, Michael Auli, and David Grangier. 2018.
\newblock \href {https://doi.org/10.18653/v1/D18-1045} {Understanding
  back-translation at scale}.
\newblock In \emph{Proceedings of the 2018 Conference on Empirical Methods in
  Natural Language Processing}, pages 489--500, Brussels, Belgium. Association
  for Computational Linguistics.

\bibitem[{Elhadad et~al.(2013)Elhadad, Miranda-Jim{\'e}nez, Steinberger, and
  Giannakopoulos}]{elhadad-etal-2013-multi}
Michael Elhadad, Sabino Miranda-Jim{\'e}nez, Josef Steinberger, and George
  Giannakopoulos. 2013.
\newblock \href {https://www.aclweb.org/anthology/W13-3102} {Multi-document
  multilingual summarization corpus preparation, part 2: {C}zech, {H}ebrew and
  {S}panish}.
\newblock In \emph{Proceedings of the {M}ulti{L}ing 2013 Workshop on
  Multilingual Multi-document Summarization}, pages 13--19, Sofia, Bulgaria.
  Association for Computational Linguistics.

\bibitem[{Gehrmann et~al.(2018)Gehrmann, Deng, and Rush}]{gehrmann2018bottom}
Sebastian Gehrmann, Yuntian Deng, and Alexander~M Rush. 2018.
\newblock Bottom-up abstractive summarization.
\newblock \emph{arXiv preprint arXiv:1808.10792}.

\bibitem[{Giannakopoulos(2013)}]{giannakopoulos-2013-multi}
George Giannakopoulos. 2013.
\newblock \href {https://www.aclweb.org/anthology/W13-3103} {Multi-document
  multilingual summarization and evaluation tracks in {ACL} 2013 {M}ulti{L}ing
  workshop}.
\newblock In \emph{Proceedings of the {M}ulti{L}ing 2013 Workshop on
  Multilingual Multi-document Summarization}, pages 20--28, Sofia, Bulgaria.
  Association for Computational Linguistics.

\bibitem[{Giannakopoulos et~al.(2015)Giannakopoulos, Kubina, Conroy,
  Steinberger, Favre, Kabadjov, Kruschwitz, and
  Poesio}]{giannakopoulos-etal-2015-multiling}
George Giannakopoulos, Jeff Kubina, John Conroy, Josef Steinberger, Benoit
  Favre, Mijail Kabadjov, Udo Kruschwitz, and Massimo Poesio. 2015.
\newblock \href {https://doi.org/10.18653/v1/W15-4638} {{M}ulti{L}ing 2015:
  Multilingual summarization of single and multi-documents, on-line fora, and
  call-center conversations}.
\newblock In \emph{Proceedings of the 16th Annual Meeting of the Special
  Interest Group on Discourse and Dialogue}, pages 270--274, Prague, Czech
  Republic. Association for Computational Linguistics.

\bibitem[{Hermann et~al.(2015)Hermann, Ko\v{c}isk\'{y}, Grefenstette, Espeholt,
  Kay, Suleyman, and Blunsom}]{10.5555/2969239.2969428}
Karl~Moritz Hermann, Tom\'{a}\v{s} Ko\v{c}isk\'{y}, Edward Grefenstette, Lasse
  Espeholt, Will Kay, Mustafa Suleyman, and Phil Blunsom. 2015.
\newblock Teaching machines to read and comprehend.
\newblock In \emph{Proceedings of the 28th International Conference on Neural
  Information Processing Systems - Volume 1}, NIPS’15, page 1693–1701,
  Cambridge, MA, USA. MIT Press.

\bibitem[{Hoang et~al.(2018)Hoang, Koehn, Haffari, and
  Cohn}]{hoang-etal-2018-iterative}
Vu~Cong~Duy Hoang, Philipp Koehn, Gholamreza Haffari, and Trevor Cohn. 2018.
\newblock \href {https://doi.org/10.18653/v1/W18-2703} {Iterative
  back-translation for neural machine translation}.
\newblock In \emph{Proceedings of the 2nd Workshop on Neural Machine
  Translation and Generation}, pages 18--24, Melbourne, Australia. Association
  for Computational Linguistics.

\bibitem[{Hu et~al.(2015)Hu, Chen, and Zhu}]{hu-etal-2015-lcsts}
Baotian Hu, Qingcai Chen, and Fangze Zhu. 2015.
\newblock \href {https://doi.org/10.18653/v1/D15-1229} {{LCSTS}: A large scale
  {C}hinese short text summarization dataset}.
\newblock In \emph{Proceedings of the 2015 Conference on Empirical Methods in
  Natural Language Processing}, pages 1967--1972, Lisbon, Portugal. Association
  for Computational Linguistics.

\bibitem[{Koupaee and Wang(2018)}]{DBLP:journals/corr/abs-1810-09305}
Mahnaz Koupaee and William~Yang Wang. 2018.
\newblock \href {http://arxiv.org/abs/1810.09305} {Wikihow: {A} large scale
  text summarization dataset}.
\newblock \emph{CoRR}, abs/1810.09305.

\bibitem[{Kubina et~al.(2013)Kubina, Conroy, and
  Schlesinger}]{kubina-etal-2013-acl}
Jeff Kubina, John Conroy, and Judith Schlesinger. 2013.
\newblock \href {https://www.aclweb.org/anthology/W13-3104} {{ACL} 2013
  {M}ulti{L}ing pilot overview}.
\newblock In \emph{Proceedings of the {M}ulti{L}ing 2013 Workshop on
  Multilingual Multi-document Summarization}, pages 29--38, Sofia, Bulgaria.
  Association for Computational Linguistics.

\bibitem[{Leuski et~al.(2003)Leuski, Lin, Zhou, Germann, Och, and
  Hovy}]{10.1145/979872.979877}
Anton Leuski, Chin-Yew Lin, Liang Zhou, Ulrich Germann, Franz~Josef Och, and
  Eduard Hovy. 2003.
\newblock \href {https://doi.org/10.1145/979872.979877} {Cross-lingual c*st*rd:
  English access to hindi information}.
\newblock \emph{ACM Transactions on Asian Language Information Processing},
  2(3):245–269.

\bibitem[{Lewis et~al.(2019{\natexlab{a}})Lewis, Liu, Goyal, Ghazvininejad,
  Mohamed, Levy, Stoyanov, and Zettlemoyer}]{Lewis2019BARTDS}
Mike Lewis, Yinhan Liu, Naman Goyal, Marjan Ghazvininejad, Abdelrahman Mohamed,
  Omer Levy, Ves Stoyanov, and Luke Zettlemoyer. 2019{\natexlab{a}}.
\newblock Bart: Denoising sequence-to-sequence pre-training for natural
  language generation, translation, and comprehension.
\newblock \emph{ArXiv}, abs/1910.13461.

\bibitem[{Lewis et~al.(2019{\natexlab{b}})Lewis, Liu, Goyal, Ghazvininejad,
  Mohamed, Levy, Stoyanov, and Zettlemoyer}]{lewis2019bart}
Mike Lewis, Yinhan Liu, Naman Goyal, Marjan Ghazvininejad, Abdelrahman Mohamed,
  Omer Levy, Ves Stoyanov, and Luke Zettlemoyer. 2019{\natexlab{b}}.
\newblock \href {http://arxiv.org/abs/1910.13461} {Bart: Denoising
  sequence-to-sequence pre-training for natural language generation,
  translation, and comprehension}.

\bibitem[{Li et~al.(2013)Li, Forascu, El-Haj, and
  Giannakopoulos}]{li-etal-2013-multi}
Lei Li, Corina Forascu, Mahmoud El-Haj, and George Giannakopoulos. 2013.
\newblock \href {https://www.aclweb.org/anthology/W13-3101} {Multi-document
  multilingual summarization corpus preparation, part 1: {A}rabic, {E}nglish,
  {G}reek, {C}hinese, {R}omanian}.
\newblock In \emph{Proceedings of the {M}ulti{L}ing 2013 Workshop on
  Multilingual Multi-document Summarization}, pages 1--12, Sofia, Bulgaria.
  Association for Computational Linguistics.

\bibitem[{Lim et~al.(2004)Lim, Kang, and Lee}]{Lim2004MultiDocumentSU}
Jung-Min Lim, In-Su Kang, and Jong-Hyeok Lee. 2004.
\newblock Multi-document summarization using cross-language texts.
\newblock In \emph{NTCIR}.

\bibitem[{Lin(2004)}]{lin-2004-rouge}
Chin-Yew Lin. 2004.
\newblock \href {https://www.aclweb.org/anthology/W04-1013} {{ROUGE}: A package
  for automatic evaluation of summaries}.
\newblock In \emph{Text Summarization Branches Out}, pages 74--81, Barcelona,
  Spain. Association for Computational Linguistics.

\bibitem[{Liu and Lapata(2019)}]{liu-lapata-2019-text}
Yang Liu and Mirella Lapata. 2019.
\newblock \href {https://doi.org/10.18653/v1/D19-1387} {Text summarization with
  pretrained encoders}.
\newblock In \emph{Proceedings of the 2019 Conference on Empirical Methods in
  Natural Language Processing and the 9th International Joint Conference on
  Natural Language Processing (EMNLP-IJCNLP)}, pages 3730--3740, Hong Kong,
  China. Association for Computational Linguistics.

\bibitem[{Liu et~al.(2020)Liu, Gu, Goyal, Li, Edunov, Ghazvininejad, Lewis, and
  Zettlemoyer}]{liu2020multilingual}
Yinhan Liu, Jiatao Gu, Naman Goyal, Xian Li, Sergey Edunov, Marjan
  Ghazvininejad, Mike Lewis, and Luke Zettlemoyer. 2020.
\newblock \href {http://arxiv.org/abs/2001.08210} {Multilingual denoising
  pre-training for neural machine translation}.

\bibitem[{Nallapati et~al.(2016)Nallapati, Zhou, dos Santos, G\.ul{\c{c}}ehre,
  and Xiang}]{nallapati-etal-2016-abstractive}
Ramesh Nallapati, Bowen Zhou, Cicero dos Santos, {\c{C}}a{\u{g}}lar
  G\.ul{\c{c}}ehre, and Bing Xiang. 2016.
\newblock \href {https://doi.org/10.18653/v1/K16-1028} {Abstractive text
  summarization using sequence-to-sequence {RNN}s and beyond}.
\newblock In \emph{Proceedings of The 20th {SIGNLL} Conference on Computational
  Natural Language Learning}, pages 280--290, Berlin, Germany. Association for
  Computational Linguistics.

\bibitem[{Narayan et~al.(2018)Narayan, Cohen, and Lapata}]{xsum-emnlp}
Shashi Narayan, Shay~B. Cohen, and Mirella Lapata. 2018.
\newblock Don't give me the details, just the summary! {T}opic-aware
  convolutional neural networks for extreme summarization.
\newblock In \emph{Proceedings of the 2018 Conference on Empirical Methods in
  Natural Language Processing}, Brussels, Belgium.

\bibitem[{Nguyen and Daum{\'e}~III(2019)}]{nguyen-daume-iii-2019-global}
Khanh Nguyen and Hal Daum{\'e}~III. 2019.
\newblock \href {https://doi.org/10.18653/v1/D19-5411} {Global voices: Crossing
  borders in automatic news summarization}.
\newblock In \emph{Proceedings of the 2nd Workshop on New Frontiers in
  Summarization}, pages 90--97, Hong Kong, China. Association for Computational
  Linguistics.

\bibitem[{Or{\u{a}}san and Chiorean(2008)}]{orasan-chiorean-2008-evaluation}
Constantin Or{\u{a}}san and Oana~Andreea Chiorean. 2008.
\newblock \href
  {http://www.lrec-conf.org/proceedings/lrec2008/pdf/539_paper.pdf} {Evaluation
  of a cross-lingual {R}omanian-{E}nglish multi-document summariser}.
\newblock In \emph{LREC 2008}.

\bibitem[{Ott et~al.(2019)Ott, Edunov, Baevski, Fan, Gross, Ng, Grangier, and
  Auli}]{ott2019fairseq}
Myle Ott, Sergey Edunov, Alexei Baevski, Angela Fan, Sam Gross, Nathan Ng,
  David Grangier, and Michael Auli. 2019.
\newblock fairseq: A fast, extensible toolkit for sequence modeling.
\newblock In \emph{Proceedings of NAACL-HLT 2019: Demonstrations}.

\bibitem[{Ouyang et~al.(2019)Ouyang, Song, and
  McKeown}]{ouyang-etal-2019-robust}
Jessica Ouyang, Boya Song, and Kathy McKeown. 2019.
\newblock \href {https://doi.org/10.18653/v1/N19-1204} {A robust abstractive
  system for cross-lingual summarization}.
\newblock In \emph{Proceedings of the 2019 Conference of the North {A}merican
  Chapter of the Association for Computational Linguistics: Human Language
  Technologies, Volume 1 (Long and Short Papers)}, pages 2025--2031,
  Minneapolis, Minnesota. Association for Computational Linguistics.

\bibitem[{Rush et~al.(2015)Rush, Chopra, and Weston}]{rush-etal-2015-neural}
Alexander~M. Rush, Sumit Chopra, and Jason Weston. 2015.
\newblock \href {https://doi.org/10.18653/v1/D15-1044} {A neural attention
  model for abstractive sentence summarization}.
\newblock In \emph{Proceedings of the 2015 Conference on Empirical Methods in
  Natural Language Processing}, pages 379--389, Lisbon, Portugal. Association
  for Computational Linguistics.

\bibitem[{Schuster et~al.(2019)Schuster, Gupta, Shah, and
  Lewis}]{Schuster2019CrosslingualTL}
Sebastian Schuster, Sonal Gupta, Rushin Shah, and Mike Lewis. 2019.
\newblock \href {https://doi.org/10.18653/v1/N19-1380} {Cross-lingual transfer
  learning for multilingual task oriented dialog}.
\newblock In \emph{Proceedings of the 2019 Conference of the North {A}merican
  Chapter of the Association for Computational Linguistics: Human Language
  Technologies, Volume 1 (Long and Short Papers)}, pages 3795--3805,
  Minneapolis, Minnesota. Association for Computational Linguistics.

\bibitem[{See et~al.(2017)See, Liu, and Manning}]{see-etal-2017-get}
Abigail See, Peter~J. Liu, and Christopher~D. Manning. 2017.
\newblock \href {https://doi.org/10.18653/v1/P17-1099} {Get to the point:
  Summarization with pointer-generator networks}.
\newblock In \emph{Proceedings of the 55th Annual Meeting of the Association
  for Computational Linguistics (Volume 1: Long Papers)}, pages 1073--1083,
  Vancouver, Canada. Association for Computational Linguistics.

\bibitem[{Sennrich et~al.(2016)Sennrich, Haddow, and
  Birch}]{sennrich-etal-2016-improving}
Rico Sennrich, Barry Haddow, and Alexandra Birch. 2016.
\newblock \href {https://doi.org/10.18653/v1/P16-1009} {Improving neural
  machine translation models with monolingual data}.
\newblock In \emph{Proceedings of the 54th Annual Meeting of the Association
  for Computational Linguistics (Volume 1: Long Papers)}, pages 86--96, Berlin,
  Germany. Association for Computational Linguistics.

\bibitem[{Shen et~al.(2018)Shen, Chen, Yang, Liu, and
  Sun}]{10.1109/TASLP.2018.2842432}
Shi-qi Shen, Yun Chen, Cheng Yang, Zhi-yuan Liu, and Mao-song Sun. 2018.
\newblock \href {https://doi.org/10.1109/TASLP.2018.2842432} {Zero-shot
  cross-lingual neural headline generation}.
\newblock \emph{IEEE/ACM Trans. Audio, Speech and Lang. Proc.},
  26(12):2319–2327.

\bibitem[{Song et~al.(2020)Song, Wang, Feng, Ren, and
  Liu}]{control-over-copying:2020}
Kaiqiang Song, Bingqing Wang, Zhe Feng, Liu Ren, and Fei Liu. 2020.
\newblock Controlling the amount of verbatim copying in abstractive
  summarization.
\newblock In \emph{Proceedings of the AAAI Conference on Artificial
  Intelligence}.

\bibitem[{Vinyals et~al.(2015)Vinyals, Fortunato, and Jaitly}]{NIPS2015_5866}
Oriol Vinyals, Meire Fortunato, and Navdeep Jaitly. 2015.
\newblock \href {http://papers.nips.cc/paper/5866-pointer-networks.pdf}
  {Pointer networks}.
\newblock In C.~Cortes, N.~D. Lawrence, D.~D. Lee, M.~Sugiyama, and R.~Garnett,
  editors, \emph{Advances in Neural Information Processing Systems 28}, pages
  2692--2700. Curran Associates, Inc.

\bibitem[{Wan(2011)}]{wan-2011-bilingual}
Xiaojun Wan. 2011.
\newblock \href {https://doi.org/10.1162/COLI_a_00061} {Bilingual co-training
  for sentiment classification of {C}hinese product reviews}.
\newblock \emph{Computational Linguistics}, 37(3):587--616.

\bibitem[{Wan et~al.(2010)Wan, Li, and Xiao}]{wan-etal-2010-cross}
Xiaojun Wan, Huiying Li, and Jianguo Xiao. 2010.
\newblock \href {https://www.aclweb.org/anthology/P10-1094} {Cross-language
  document summarization based on machine translation quality prediction}.
\newblock In \emph{Proceedings of the 48th Annual Meeting of the Association
  for Computational Linguistics}, pages 917--926, Uppsala, Sweden. Association
  for Computational Linguistics.

\bibitem[{Wan et~al.(2019)Wan, Luo, Sun, Huang, and Yao}]{wan2019cross}
Xiaojun Wan, Fuli Luo, Xue Sun, Songfang Huang, and Jin-ge Yao. 2019.
\newblock Cross-language document summarization via extraction and ranking of
  multiple summaries.
\newblock \emph{Knowledge and Information Systems}, 58(2):481--499.

\bibitem[{Yao et~al.(2015)Yao, Wan, and Xiao}]{yao-etal-2015-phrase}
Jin-ge Yao, Xiaojun Wan, and Jianguo Xiao. 2015.
\newblock \href {https://doi.org/10.18653/v1/D15-1012} {Phrase-based
  compressive cross-language summarization}.
\newblock In \emph{Proceedings of the 2015 Conference on Empirical Methods in
  Natural Language Processing}, pages 118--127, Lisbon, Portugal. Association
  for Computational Linguistics.

\bibitem[{{Zhang} et~al.(2016){Zhang}, {Zhou}, and {Zong}}]{7502066}
J.~{Zhang}, Y.~{Zhou}, and C.~{Zong}. 2016.
\newblock Abstractive cross-language summarization via translation model
  enhanced predicate argument structure fusing.
\newblock \emph{IEEE/ACM Transactions on Audio, Speech, and Language
  Processing}, 24(10):1842--1853.

\bibitem[{Zhu et~al.(2018)Zhu, Li, Liu, Zhou, Zhang, and
  Zong}]{zhu-etal-2018-msmo}
Junnan Zhu, Haoran Li, Tianshang Liu, Yu~Zhou, Jiajun Zhang, and Chengqing
  Zong. 2018.
\newblock \href {https://doi.org/10.18653/v1/D18-1448} {{MSMO}: Multimodal
  summarization with multimodal output}.
\newblock In \emph{Proceedings of the 2018 Conference on Empirical Methods in
  Natural Language Processing}, pages 4154--4164, Brussels, Belgium.
  Association for Computational Linguistics.

\bibitem[{Zhu et~al.(2019)Zhu, Wang, Wang, Zhou, Zhang, Wang, and
  Zong}]{zhu-etal-2019-ncls}
Junnan Zhu, Qian Wang, Yining Wang, Yu~Zhou, Jiajun Zhang, Shaonan Wang, and
  Chengqing Zong. 2019.
\newblock \href {https://doi.org/10.18653/v1/D19-1302} {{NCLS}: Neural
  cross-lingual summarization}.
\newblock In \emph{Proceedings of the 2019 Conference on Empirical Methods in
  Natural Language Processing and the 9th International Joint Conference on
  Natural Language Processing (EMNLP-IJCNLP)}, pages 3054--3064, Hong Kong,
  China. Association for Computational Linguistics.

\end{thebibliography}

\clearpage
\newpage
\appendix
\section{Appendix}

\begin{table*}[t]
\centering
\setlength{\tabcolsep}{3pt}
\scalebox{0.9}{
\begin{tabular}{p{15cm}}
\toprule
\textbf{Topic:} How to Reduce the Redness of Sunburn: Healing and Concealing Sunburns. \\
\toprule
\textbf{Article:} Try to drink at least 10 full glasses of water each day for a week after your sunburn ... This is the traditional go-to remedy when dealing with a burn. The gel of the aloe vera plant has natural anti-inflammatory properties and can speed up the healing process if applied correctly ... Get out a small bowl and mix equal parts baking soda and cornstarch ... You can use the leaves and bark of the witch hazel plant for medicinal purposes ... You can fill up a bottle and spray the vinegar directly on your skin for relief ... Many natural healers swear that potatoes can reduce pain and inflammation. Get a few potatoes and use a knife to cut them into thin slices ... This one is a bit of a long-shot but, if nothing else, the cool temperature of the yogurt may soothe your skin ... Light, cotton garments that fall away from the skin are your best options during your recovery period ... Apply a green-tinted primer to the burned areas to counterbalance the appearance of redness  ... 
  \\ 
\toprule
\textbf{Reference:} Drink a lot of water. Apply aloe vera. Create a baking soda paste. Use witch hazel. Apply apple cider vinegar to the area. Apply potato slices to the area. Apply live cultured yogurt. Wear loose and dark clothing. Use make-up to cover the redness.
 \\ 
\toprule
\textbf{Trans-Sum:} Drink plenty of water. Apply aloe vera gel to the skin. Make a baking salt and corn flour mask. Use hazelnut extract. Apply apple cider vinegar. Use potatoes. Apply yogurt to the skin. Apply blush.
 \\
\toprule
\textbf{Trans-Sum-R:} Drink plenty of water. Apply aloe vera gel. Use baking salt and cornmeal. Use hazelnuts and bark. Apply apple cider vinegar. Apply potatoes. Apply yogurt to the skin. Avoid wearing makeup.
 \\ 
\toprule
\textbf{DC+Synth+MT (Ours):}  Drink plenty of water. Apply aloe vera gel to the burn. Mix baking soda and cornstarch. Use witch hazel. Apply apple cider vinegar to the burn. Use potato slices. Apply yogurt to the burn. Wear dark clothing.
 \\
\toprule
\end{tabular}
}
\caption{An example output summary for Trans-Sum, Trans-Sum-R and DC+Synth+MT. Human annotators preferred the output from DC+Synth+MT.}
\label{tab:example_appendix_1}
\end{table*}

\begin{table*}[t]
\centering
\setlength{\tabcolsep}{3pt}
\scalebox{0.9}{
\begin{tabular}{p{15cm}}
\toprule
\textbf{Topic:} How to Speak Portuguese: Studying Basic Portuguese Vocabulary. \\
\toprule
\textbf{Article:}   Practice saying ordinary phrases so you can carry on a casual conversation in any situation ... Practice polite phrases like ``please'' and ``thank you''. Learn a few ways to communicate your understanding or confusion ... If you will be traveling in Portugal, learn some of the basic questions and statements you will need to get around ...  Learn how to ask for rooms, order off a menu, and talk to shopkeepers ...  Portuguese numbers sometimes reflect the gender of the word they are applied to. For instance, if you are talking about a woman, you would say ``uma mulher'', but for a man you would say ``um homem.''.
  \\ 
\toprule
\textbf{Reference:}  Learn a few greetings. Master some basic conversational phrases. Learn key travel phrases. Practice your shopping and restaurant vocabulary. Learn how to count in Portuguese.
 \\  
\toprule
\textbf{Trans-Sum:}  Learn some basic phrases to communicate basic conversation. Practice basic phrases to communicate understanding or confusion. Learn some basic questions and statements to communicate when traveling. Learn how to order a room or talk to merchants. Learn how to talk about gender.
 \\
\toprule
\textbf{Trans-Sum-R:} Learn basic conversational phrases. Use simple phrases to communicate understanding or confusion. Learn basic questions and statements when traveling to Portugal. Learn how to order a room, menu, or speak to merchants. Learn how to say “um” or “um homem” if you’re talking about a woman.
 \\ 
\toprule
\textbf{DC+Synth+MT (Ours):} Learn some basic conversational phrases. Learn some polite phrases Learn some basic phrases when communicating in public. Learn some basic phrases when communicating in public. Learn how to communicate with people. Learn how to communicate with numbers.
 \\
\toprule
\end{tabular}
}
\caption{An example output summary for Trans-Sum, Trans-Sum-R and DC+Synth+MT. Human annotators preferred the output from DC+Synth+MT.}
\label{tab:example_appendix_2}
\end{table*}

\begin{table*}[t]
\centering
\setlength{\tabcolsep}{3pt}
\scalebox{0.9}{
\begin{tabular}{p{15cm}}
\toprule
\textbf{Topic:}  How to Teach English As a Second Language to Beginners: Embracing Best Practices. \\
\toprule
\textbf{Article:}  One great way to facilitate learning is to encourage students to avoid speaking languages other than English in the classroom ... When explaining an activity or giving directions about homework, classwork, or a project, you should always give both verbal and written instructions ... This will aid in word association and in pronunciation ... No matter what type of lesson you are teaching or what activity your students are doing, you should monitor them constantly ... Teaching English as a second language to beginners is a lot more effective when you use a variety of types of learning ... When teaching beginners or very young students, break the lesson into several pieces of about 10 minutes. 
  \\ 
\toprule
\textbf{Reference:}   Encourage students to speak only English in the classroom. Provide verbal and written instructions. Monitor students’ progress constantly. Promote a diversity of modes of learning. Break lessons into small pieces.
\\  
\toprule
\textbf{Trans-Sum:}  Encourage students to speak English. Give both oral and written instructions. Control your students. Encourage different types of learning. Divide lessons into small pieces. Change your lesson types often.
\\
\toprule
\textbf{Trans-Sum-R:} Encourage students to speak English. Provide both oral and written instructions. Monitor your students. Diversify your teaching methods. Divide the lesson into short pieces. Switch up your teaching style.
 \\ 
\toprule
\textbf{DC+Synth+MT (Ours):} Encourage students to speak English. Give both verbal and written instructions. Check on students regularly. Encourage a variety of learning methods. Break your lessons down into small chunks. Vary your lesson types.
 \\ 
\toprule
\end{tabular}
}
\caption{An example output summary for Trans-Sum, Trans-Sum-R and DC+Synth+MT. Human annotators preferred the output from Trans-Sum-R and Trans-Sum over DC+Synth+MT.}
\label{tab:example_appendix_3}
\end{table*}

\begin{table*}[t]
\centering
\setlength{\tabcolsep}{3pt}
\scalebox{0.9}{
\begin{tabular}{p{15cm}}
\toprule
\textbf{Topic:} How to Live an Active Life with COPD: Participating in Exercise and Activities with COPD. \\
\toprule
\textbf{Article:}  With a serious lung disease like COPD, you have to be exceptionally careful when you start physical activity. Although exercise can help improve your COPD, you still need to ease into activities slowly ... Increasing your lifestyle activity is a great way to stay active without overdoing it. These are not cardio activities, but they also help keep your body moving and your lungs working ... When you're ready to progress to more structured exercise, you need to plan to include a warm-up. This is an essential component of safe exercise for those with COPD ... Unless cleared by your physician, you should only participate in aerobic activities that are low in intensity. This level is the most safe for patients with COPD ... Aerobic exercises are great to help improve the condition of your lungs and improve your cardiovascular system; however, strength training is an essential form of exercise as well. 
  \\ 
\toprule
\textbf{Reference:}  Ease into activities. Increase your lifestyle activity. Always do a warm-up. Add in low-intensity cardio exercises. Do light strength training. Try pilates and yoga for breathing exercises.
\\  
\toprule
\textbf{Trans-Sum:} Start slowly. Include daily activities. Include a warm-up. Perform low-intensity aerobic exercises. Perform strength training. Do yoga or pilates.
\\
\toprule
\textbf{Trans-Sum-R:} Start slowly. Increase the frequency and duration of daily activities. Warm up. Perform low-intensity aerobic exercises. Perform strength training. Do yoga or pilates.
 \\ 
\toprule
\textbf{DC+Synth+MT (Ours):} Start slowly. Include daily activities. Warm up. Do low-intensity aerobic exercise. Strength train. Do yoga or pilates.
 \\ 
\toprule
\end{tabular}
}
\caption{An example output summary for Trans-Sum, Trans-Sum-R and DC+Synth+MT. Human annotators preferred the output from Trans-Sum-R and Trans-Sum over DC+Synth+MT.}
\label{tab:example_appendix_4}
\end{table*}

\subsection{Reproducibility}\label{reproducibility}
We use Fairseq \cite{ott2019fairseq} for all our experiments. We follow the hyperparmeter settings used by \newcite{lewis2019bart} for all summarization and translation models we train.\footnote{\href{https://github.com/pytorch/fairseq/blob/master/examples/bart/README.summarization.md}{Link to hyper-parameter settings used \newcite{lewis2019bart}}.} We note that we had to make some modifications to existing mBART code, to support monolingual summarization. We will make this code, along with our data pre-processing scripts, available upon acceptance. We train all our models on a single machine with four Nvidia Tesla V100 GPUs, 96 CPU cores, and 693 GB of RAM. We train all models until the validation loss no longer improves for two epochs, and use the checkpoint with the best validation loss for inference. The average runtime for each of our training runs was between three to six hours, depending on the dataset size (it was quickest for Turkish and slowest for Spanish). 

All models that we report in Table \ref{tab:results} were trained using the exact same pre-trained mBART architecture ($\sim$ 680M parameters), with the same hyperparameters. For inference, we used a beam-size of five for all models. The ROUGE \cite{lin-2004-rouge} scores were computed using the official ROUGE script. \footnote{The parameters used to compute the ROUGE scores were ``-c 95 -r 1000 -n 2 -a".}

\subsection{Splitting English Data}\label{en_split}
To get a fair assessment of cross-lingual performance, we need to ensure, at a minimum, that any English article that is parallel to any test article in any of the four languages, gets mapped to the English test set. We note, however, that this is not sufficient, since there are multiple methods (articles) for each topic, and there may be some content overlap between them. Therefore, in addition to parallel articles, we also include all English articles that overlap in topic with any test article in any of the four non-English languages in the test set for English. While this way of splitting the data means we have fewer English articles for training, we opted for this as it ensures purity of the tests sets. Furthermore, it also ensures that models that learn topic-specific information will not be able to generalize to the test set, since there is minimal topical overlap. This method of splitting filtered out $\sim$ 72K examples to the test set, and left us with $\sim$ 69K examples for training and development sets.

\subsection{Data Statements}\label{data_statements}
All of the data was collected according with the terms and conditions listed on the website. We followed WikiHow's rate limit (four second delay between each request) while scraping the website. We follow the guidelines suggested by \newcite{bender-friedman-2018-data} and prepare a data statement, to the best of our ability, for the data we collect. 
\subsubsection{Curation Rationale}
This dataset was collected in order to enable further research into cross-lingual and multilingual summarization. We first collected English articles from WikiHow. Each English article links to any corresponding articles that may be available in the other $17$ languages that are supported on WikiHow. We use this information to collect parallel articles between English and each of the other $17$ languages. We then align these articles using the illustrative images for each of the steps detailed in the article, since these images are unique to a given step. 

\subsubsection{Language Variety}
The dataset includes articles in $18$ languages (i.e. English, Spanish, Portuguese, French,
German, 
Russian,
Italian,
Indonesian, 
Dutch,
Arabic, 
Chinese,
Vietnamese,  
Thai,
Japanese,
Korean,
Hindi,  
Czech, 
Turkish). The information about the varieties for the languages is not available.

\subsubsection{Speaker Demographic}
We do not have access to the demographics of the writers and editors of the articles.  

\subsubsection{Annotator Demographic}
We do not collect any additional annotations for this dataset.

\subsubsection{Speech Situation}
The articles written on the website are a collaborative effort from people all over the world. Each article and summary is written and edited by 23 people, and further re-viewed by 16 people, on average, which ensures that the content is of a high-quality. A majority of the non-English articles are written by people who are fluent in both English and the target language, and are further reviewed by WikiHow's international translation team, before they are published. The articles are written as how-to guides over a wide variety of topics, and the intended audience is anyone that is interested in instructions to complete a certain task.

\subsubsection{Text Characteristics}
The articles cover $19$ broad categories including health, arts and entertainment, personal care and style, travel, education and communications, etc. The categories covered a broad set of genres and topics.

\subsection{Examples}\label{appendix_examples}
We present four additional example outputs for each of the three systems that were evaluated by human annotators. We show two examples where our system (DC+Synth+MT) was preferred, in Tables \ref{tab:example_appendix_1} and \ref{tab:example_appendix_2}, and two examples where the baselines were preferred over our system, in Tables \ref{tab:example_appendix_3} and \ref{tab:example_appendix_4}. We will make the model outputs available for all systems.

\end{document}